\documentclass[sigconf]{acmart}

\copyrightyear{2024}
\acmYear{2024}
\setcopyright{rightsretained}
\acmConference[HRI '24]{Proceedings of the 2024 ACM/IEEE International Conference on Human-Robot Interaction}{March 11--14, 2024}{Boulder, CO, USA}
\acmBooktitle{Proceedings of the 2024 ACM/IEEE International Conference on Human-Robot Interaction (HRI '24), March 11--14, 2024, Boulder, CO, USA}
\acmDOI{10.1145/3610977.3634930}
\acmISBN{979-8-4007-0322-5/24/03}

\usepackage{graphicx}
\usepackage[ruled,noend,linesnumbered]{algorithm2e}

\usepackage{wrapfig}
\usepackage{amsmath}
\DeclareMathOperator*{\argmax}{arg\,max}

\usepackage{color}

\newcommand{\st}{\ensuremath{s}}

\newcommand{\ac}{\ensuremath{a}}
\newcommand{\St}{\ensuremath{\mathcal{S}}}
\newcommand{\Ac}{\ensuremath{\mathcal{A}}}

\newcommand{\reward}{\ensuremath{\mathcal{R}}}
\newcommand{\policy}{\ensuremath{\pi}}
\newcommand{\weightp}{\ensuremath{\psi}}
\newcommand{\weighta}{\ensuremath{\theta}}
\newcommand{\Weighta}{\ensuremath{\Theta}}
\newcommand{\absfunc}{\ensuremath{\hat{f}}}

\newcommand{\traj}{\ensuremath{\tau}}
\newcommand{\lang}{\ensuremath{\ell}}

\newcommand{\Capt}{\ensuremath{C}}
\newcommand{\feat}{\ensuremath{\phi}}

\newcommand\figref[1]{Fig.~\ref{#1}}
\newcommand\equref[1]{Eq.~\eqref{#1}}
\newcommand\secref[1]{Sec.~\ref{#1}}

\begin{document}

\title{Preference-Conditioned Language-Guided Abstraction}

\author{Andi Peng}
\affiliation{
  \institution{MIT}
  \country{United States of America}
}
\author{Andreea Bobu}
\affiliation{
  \institution{Boston Dynamics AI Institute}
  \country{United States of America}
}
\author{Belinda Z. Li}
\affiliation{
  \institution{MIT}
  \country{United States of America}
}
\author{Theodore R. Sumers}
\affiliation{
  \institution{Princeton}
  \country{United States of America}
}
\author{Ilia Sucholutsky}
\affiliation{
  \institution{Princeton}
  \country{United States of America}
}
\author{Nishanth Kumar}
\affiliation{
  \institution{MIT}
  \country{United States of America}
}
\author{Thomas L. Griffiths}
\affiliation{
  \institution{Princeton}
  \country{United States of America}
}
\author{Julie A. Shah}
\affiliation{
  \institution{MIT}
  \country{United States of America}
}
\renewcommand{\shortauthors}{Andi Peng et al.}

\begin{abstract}
Learning from demonstrations is a common way for users to teach robots, but it is prone to spurious feature correlations. Recent work constructs \textit{state abstractions}, i.e. visual representations containing task-relevant features, from language as a way to perform more generalizable learning.
However, these abstractions also depend on a user's \textit{preference} for what matters in a task, which may be hard to describe or infeasible to exhaustively specify using language alone.
How do we construct abstractions to capture these latent preferences? We observe that how humans behave reveals how they see the world.
Our key insight is that changes in human behavior inform us that there are differences in preferences for how humans see the world, i.e. their state abstractions.
In this work, we propose using language models (LMs) to query for those preferences directly given knowledge that a change in behavior has occurred.
In our framework, we use the LM in two ways: first, given a text description of the task and knowledge of behavioral change between states, we query the LM for possible hidden preferences; second, given the most likely preference, we query the LM to construct the state abstraction. In this framework, the LM is also able to \textit{ask the human directly} when uncertain about its own estimate.
We demonstrate our framework's ability to construct effective preference-conditioned abstractions in simulated experiments, a user study, as well as on a real Spot robot performing mobile manipulation tasks.
\end{abstract}



%\keywords{state abstraction, learning from human input, human preferences}

\maketitle

\section{Introduction}
\label{sec:intro}

\begin{figure*}
    \centering
    \includegraphics[width=\textwidth,trim={0 0 0 0},clip]{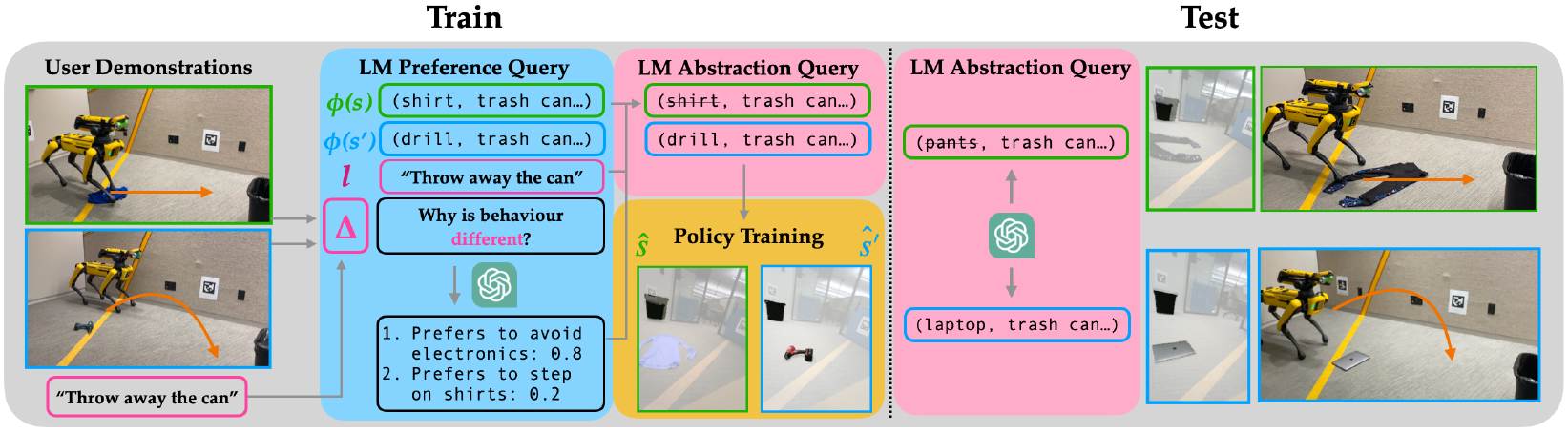}
    \vspace{-4mm}
    \caption{Preference-Conditioned Language-Guided Abstraction (PLGA). (Left) The robot uses the demonstration pair to identify a behavior change not captured by the language specification. Given this information, we query the LM for potential preferences that could explain this change. Finally, the robot uses its best preference estimate to query the LM for state abstractions and train a policy. (Right) At test time, the robot generalizes to new states and language specifications using its preference-conditioned abstractions.}
\label{fig:front_fig}
\end{figure*}

In robot learning, we wish to teach robots how to perform tasks that human users want.
Learning from demonstrations (LfD) is a common way for doing so, as the user can directly teach the robot desired task behavior. Unfortunately, LfD requires a lot of data and often fails to fully specify all the reasons behind the demonstrated behavior \cite{correa2022humans}. For example, consider the scenario depicted in \figref{fig:front_fig}, which shows two demonstrations for the task ``throw away the can''. Is the user demonstrating moving cans, navigating to a specific goal location, or tossing the can in the trash? Without more data disambiguating the demonstrations, it's difficult for the robot to fully learn what all the features that matter for the task are.

Humans, meanwhile, exhibit extraordinary generalization capabilities in new environments. A key reason why humans can learn so quickly is their ability to construct simplified mental representations over which to plan \cite{ho2022people}. Useful abstractions are task-dependent, and prior experience, commonsense reasoning, and direct teaching contribute to humans learning how to best construct these abstractions \cite{ho2023rational,huey2023visual}. Recent work showed how we can successfully leverage strong priors embedded in LMs to aide in constructing state abstractions for robots \cite{peng2023lga}. Given a language description of the task, language-guided abstraction (LGA) leverages the strong semantic priors in LMs to model task-relevant features important for decision-making \cite{peng2023lga}.

Unfortunately, LGA is limited when the features that are important to the human are not \textit{fully} specified in language. This presents a challenge in real-world robotics settings where we must adapt to human preferences quickly and efficiently, which can often be expensive or even intractable for preferences inexpressible through natural language. How can we ensure that the robot's state abstractions are strong enough to enable efficient learning \cite{peng2023diagnosis,peng2023lga} yet flexible enough to learn individual preferences? 

In this work, we propose a framework to use language and behavior to query LMs for their possible abstraction preference. Our observation is how humans behave is indicative of how they see the world, i.e. their state abstraction. If we are able to observe a difference in human behavior, this provides meaningful grounds to infer there are differences in preferences for how their abstractions are constructed. In this work, we introduce Preference-conditioned Language-Guided Abstraction (PLGA), a framework for using this information to infer latent preferences to explain differences in human behavior. In PLGA, we use the LM in two ways: first, given a text description of the task and knowledge of behavior change between states, we query the LM for possible hidden preferences; second, given the most likely preference, we query the LM for the state abstraction. In this framework, the LM is also able to \textit{actively query for human preferences} by asking the human when it is uncertain about its own estimate. With these preferences, we construct preference-conditioned abstractions for downstream learning.

The roadmap for this paper is as follows: In \secref{sec:problem} we formalize our problem formulation and the task of learning preference-conditioned state abstractions. In \secref{sec:method} we describe two versions of our method (PLGA): in passive PLGA, we use LMs to ``simulate'' human preferences; in active PLGA, we (may) additionally query humans for their preferences. We then conduct several experiments that demonstrate the effectiveness of passive PLGA (\secref{sec:sim-exps}) and active PLGA (\secref{sec:human-exps}) in simulated environments, and passive PLGA in a real-world robotics environment (\secref{sec:robot-exps}). In all settings, we find that PLGA is able to successfully capture human preferences, producing state abstractions that enable generalization across tasks, while also improving the user interaction experience beyond LGA.
% \vspace{-4mm}
\section{Problem Formulation}
\label{sec:problem}

\subsection{Preliminaries}

\noindent\textit{Markov Decision Processes.}
We model our problem as a Markov Decision Process $\mathcal{M} = \langle \St, \Ac, \mathcal{T}, \reward \rangle$ with states $\st\in\St$, actions $\ac\in\Ac$, transition probability $\mathcal{T}:\St \times \Ac \times \St \rightarrow [0,1]$, and rewards $\reward:\St \times \Ac \rightarrow \mathbb{R}$. We define a trajectory $\traj$ as a sequence of state-action pairs, $\traj = (\st_0, \ac_0,\cdots,\st_T, \ac_T)$. We wish to learn a policy $\policy_\weightp:\St \rightarrow \Ac$, parameterized by $\weightp$, that solves the MDP.

\smallskip
\noindent\textit{Goal-Conditioned Behavioral Cloning.}
We consider scenarios where the robot does not know the reward, and instead it learns the policy $\policy_\weightp$ from user demonstrations $\mathcal{D} = \{\traj^i\}_{i=1}^n = \{(\st_1^i, \ac_1^i, ..., \st_T^i, \ac_T^i)\}_{i=1}^{n}$ and a natural language description $\lang \in \mathcal{L}$ that specifies the goal for each demonstration. Goal-conditioned behavioral cloning (GCBC) \cite{co2018guiding} is a method where the policy can condition on both the current state $s$ and a linguistically-specified goal $\lang$ to try and imitate human actions. GCBC attempts to learn a policy $\pi$ that minimizes:

\vspace{-1em}
\begin{equation}
\begin{aligned}
\mathcal{L}_{\mathrm{GCBC}} = 
\mathbb{E}_{(\st_t^i,\ac_t^i,\lang^i)\sim D_{\mathrm{train}}}[\|\policy_{\weightp}(\st_t^i, \lang^i)-a^i_t\|^2_2]\enspace,
\end{aligned}
\label{eq_gcbc}
\end{equation}

However, because at its core the algorithm simply imitates the data it has seen, GCBC alone cannot reliably generalize the policy $\policy_\weightp(\st^i_t, \lang^i)$ to novel specifications $\lang^i$ or states $s^i_t$.

\smallskip
\noindent\textit{Language-Guided Abstraction.}
Our work builds on LGA (Language-Guided Abstraction)~\cite{peng2023lga}, which proposes using LM priors to build abstract state representations. 
LGA's key novelty is an abstraction function $\absfunc: \St \times \mathcal{L} \to \hat{\St}$ that contextualizes the state within the language task specification and produces a task-relevant state abstraction $\hat\st = \absfunc(\st, \lang)$.
This extends GCBC to learning policies $\policy_{\hat\weightp}: \hat{\St} \to \Ac$ that operate at the abstraction level:

\begin{equation}
\begin{aligned}
    \mathcal{L}_{\mathrm{LGA}} = 
   \mathbb{E}_{(\st_t^i,\ac_t^i,\lang^i)\sim \mathcal{D}}
   [||\textcolor{red}{\policy_{\hat{\weightp}}}(\mathbin{\textcolor{red}{\absfunc(\st_t^i, \lang^i)}})-\ac^i_t||^2_2]\enspace.
\end{aligned}
\label{eq_lga}
\end{equation}

The key to LGA generalizing beyond specific user commands and demonstrations is the rich language prior that determines which states and specifications should be treated similarly in the context of decision-making (e.g. if the robot has learned to ``pick up a cup'', it should also know to ``pick up something to drink with''). 

In LGA, the abstraction function $\absfunc^{\mathrm{LGA}}$ consists of 3 steps:
\begin{enumerate}
    \item In \textbf{textualization}, a state captioner $\Capt: \St \to \mathcal{L}^\phi$ converts the raw perceptual state $\st$ into a text-based feature set $\feat=\Capt(\st)$. This text representation may include common visual attributes of the state like object type and color, which are reasonably accessible via segmentation models today \cite{kirillov2023segment}. 
    \item \textbf{Feature abstraction} passes $\feat$ and $\lang$ to the LM and asks for the features relevant for the task, $\hat\feat = \mathrm{LM}_\mathrm{abs}(\feat, \lang)$. We denote $\mathrm{LM}_\mathrm{abs}$ as queries for the abstraction, e.g. ``What features in the scene matter for the task $\langle$throw away the can$\rangle$?''.
    \item Lastly, LGA \textbf{instantiates} $\hat\feat$ into an abstracted state $\hat\st = \Capt^{-1}(\hat\feat)$. We assume that the captioner from step 1 is invertible and can, thus, instantiate (potentially abstracted) perceptual states from feature sets, i.e. $\Capt^{-1}: \mathcal{L}^\phi \to \St$. For instance, in \figref{fig:front_fig} the captioner converts states to a feature set of object names, and the inverse captioner takes an LM-obtained feature set and converts it into an abstracted state.
\end{enumerate}
Altogether, the LGA abstraction function can be written as $\absfunc^{\mathrm{LGA}}(\st, \lang) = \Capt^{-1}(\mathrm{LM}_\mathrm{abs}(\Capt(\st), \lang))$.  

\subsection{Problem Statement}
Unfortunately, LGA is limited when the language utterance does not fully specify the desired behavior. For example, in \figref{fig:front_fig}, without explicitly mentioning ``avoid electronics'' in the utterance $\lang$, there is no recourse for the model to know that ``drill'' or ``laptop'' should be captured by the abstraction, and are thus relevant for robot behavior. Consequently, the LGA function $\absfunc$ will ignore it, leading to learning an incorrect policy $\policy_{\hat\weightp}$ downstream.
In this paper, we present a method to infer and incorporate such unexpressed preferences.

Formally, we assume the human holds a latent preference $\weighta\in\Weighta$ over what the abstraction $\hat\st$ should be, i.e. $\hat\st = \absfunc(\st, \lang, \weighta)$ for $\absfunc: \St \times \mathcal{L} \times \Theta \to \hat{\St}$. In the example above, the user is a cautious person who prefers to ``avoid electronics''. The challenge is that the robot does not know $\weighta$ and must infer it in order to build the abstraction.

We observe that in providing demonstrations to the robot, humans reveal information about what matters to them in their tasks. In other words, demonstrations \textit{implicitly} give evidence for what the latent abstraction preference $\weighta$ is~\citep{ jeon2020rewardrational}. In this paper, we study how we can use demonstrations $\mathcal{D}$ together with the utterance $\lang$ to learn \textit{preference-conditioned} language-guided abstractions $\hat\st = \absfunc(\st, \lang, \weighta)$, i.e. abstractions that capture \emph{how the human} represents the task, using information from both their linguistic specification and physical behaviors. We expect these preference-conditioned abstractions will allow flexible adaptation to preferences over task completion.
\section{Method: Preference-conditioned Language-Guided Abstraction}
\label{sec:method}

We present our method for constructing preference-conditioned language guided abstractions (PLGA). We use an LM to give a common-sense prior over abstraction preferences given a language specification and information about user demonstrations. At a high level, our method consists of two steps: 1) estimating the abstraction preference $\weighta$ and 2) updating the abstraction function $\absfunc$ with that $\weighta$.
Our use of the LM is, thus, two-fold: first, given $\lang$ and information about demonstrations $\tau$, we query the LM for most likely human preference $\weighta$; next, given that preference, we query the LM for the abstraction. This framing also allows us to actively query the human for their preference when the LM is uncertain about its set of hypothesized $\weighta$s. We present the full PLGA procedure in Alg. \ref{alg:PLGA}.

We use GPT4~\cite{openai2023gpt4} as our LM to query for human preferences and state abstractions given state, language, and trajectory information. Here, we first focus on LM queries for state abstractions. We discuss the use of LMs for querying for human preferences in \secref{sec:lm_pref}.

\begin{algorithm}
\textbf{Input:} $N$ sampled trajectory pairs $(\tau, \tau') \in \mathcal{D}$, specification $\lang$, captioner $\Capt$, entropy threshold $\epsilon$, distance threshold $\kappa$ \\

\textbf{Init:} Abstraction model without preferences $\absfunc^{\mathrm{LGA}}$\\
\For{$i\leftarrow 1$ \KwTo $N$}{
\textcolor{orange}{/\ /\ Language can't explain behavior change} \\
\If{$\|\traj - \traj'\|^2_2 > \kappa$ and $\absfunc^{\mathrm{LGA}}(\st, \lang) = \absfunc^{\mathrm{LGA}}(\st', \lang)$}{
\textcolor{orange}{/\ /\ Find hidden preference as in \secref{sec:lm_pref}} \\
$\Weighta_{LM}, P(\weighta \mid \st, \st', \lang, \Delta=1) \sim \mathrm{LM}_\mathrm{pref}(\Capt(\st), \Capt(\st'), \lang, \Delta=1)$ \\
\textcolor{orange}{/\ /\ LM is confident about preference}\\
\If{$H(P(\weighta \mid \st, \st', \lang, \Delta=1)) < \epsilon$}{
    $\hat\weighta \gets \argmax_\weighta(P(\weighta \mid \st, \st', \lang, \Delta=1))$ \\
}
\Else{
    $\hat\weighta \gets$ query H \textcolor{orange}{/\ /\ as in \secref{sec:lm_active}}
}
}
}

\textcolor{orange}{/\ /\ Create updated abstractions as in \secref{sec:lm_abs}.} \\
$\absfunc^{\mathrm{PLGA}}(\st, \lang, \hat\weighta) = \Capt^{-1}(\mathrm{LM}_\mathrm{abs}(\Capt(\st), \lang, \hat\weighta))$ \\

$\policy_{\hat\weightp} \gets \mathcal{L}_{\mathrm{PLGA}} (\absfunc^{PLGA}(\st, \lang, \hat\weighta))$ 

\caption{PLGA}
\label{alg:PLGA}
\end{algorithm}

\subsection{LMs as Models of State Abstraction}
\label{sec:lm_abs}

Moving beyond LGA, we want an abstraction function that is preference-conditioned. Here, we assume we already have an estimate of the human's abstraction preference $\weighta$, and we discuss the estimation process later in \secref{sec:lm_pref}.
We can use the same captioner from LGA, but the LM must now be queried with preference information as well. Hence, in our feature abstraction step we pass $\feat$, $\lang$ and a language description of the estimate $\weighta$ to the LM and query it for the preference-conditioned features that are relevant for the task, i.e. $\hat\feat = \mathrm{LM}_\mathrm{abs}(\feat, \lang, \weighta)$. In the \figref{fig:front_fig} example, the abstraction query includes not only the scene and task specification, but also the inferred preference ``avoid electronics''. Overall, our abstraction function can be written as $\absfunc^{\mathrm{PLGA}}(\st, \lang, \weighta) = \Capt^{-1}(\mathrm{LM}_\mathrm{abs}(\Capt(\st), \lang, \weighta))$. 

\smallskip
\noindent\textbf{Probabilistic Interpretation.} 
Given the state $\st$ and language specification $\lang$ only, we would ideally like a model $P(\hat\st \mid \st, \lang)$ that specifies what the abstracted state $\hat\st$ should be. At its core, LGA leverages the LM's  prior to model this probability, querying the LM for the most likely abstraction, i.e. $\absfunc^\mathrm{LGA}(\st, \lang) = \argmax_{\hat\st} P(\hat\st \mid \st, \lang)$.

In real-world settings, however, $\st$ and $\lang$ may not contain sufficient information for the LM to accurately approximate the abstraction. 
Rather, there is an additional dependency on the (latent) abstraction \emph{preference} $\weighta$, which gives $P(\hat\st \mid \st, \lang) = \sum_{\weighta\in\Weighta} P(\hat\st \mid \st, \lang, \weighta) P(\weighta \mid \st, \lang)$.
Instead of computing the full sum, we simply estimate the most likely $\weighta$ in \secref{sec:lm_pref}, then use it in $P(\hat\st \mid \st, \lang, \weighta)$. If we already have an estimate $\weighta$, PLGA assumes the LM has a strong prior for modeling $P(\hat\st \mid \st, \lang, \weighta)$ and we can query the LM for the most likely abstraction, i.e. $\absfunc^\mathrm{PLGA}(\st, \lang, \weighta) = \argmax_{\hat\st} P(\hat\st \mid \st, \lang, \weighta)$.

% \vspace{-2mm}
\subsection{LMs as Models of Preference}
\label{sec:lm_pref}

We now discuss how PLGA estimates the human's latent abstraction preference parameter $\weighta$. Given $\st$ and $\lang$, we could query an LM for potential human preferences $\weighta_i$ corresponding to that state and task specification, i.e. $\weighta_i \sim \mathrm{LM}_\mathrm{pref}(\Capt(\st), \lang)$, but the space of possible preferences may be intractably large. For example, in \figref{fig:front_fig} the more objects in the scene, the combinatorially more preferences for caring or not caring about each one of them the LM could find. 

We observe that given demonstrations $\tau$, we can derive additional insights about the abstraction preference beyond the language specification: human behavior (i.e. demonstrations) implicitly reveals information about what the human cares about in the world (i.e. the abstraction). If we had a language description of the demonstrations, we could include it in our query to the LM. Unfortunately, behaviors are particularly challenging to caption~\cite{rana2023behaviorcaptioning} and asking the human to narrate every demonstration they give is too burdensome.

Instead of giving the LM a description of the behavior the human demonstrates, we indicate initial scenes where behaviors are \textit{different} in ways that the language utterance does not specify. 
Given a trajectory pair $(\traj, \traj')$ corresponding to initial states $\st$ and $\st'$ and the specification $\lang$, we introduce a binary variable $\Delta(\st, \st', \lang)$ that indicates whether the desired human behaviors in $\st$ and $\st'$ are different in ways not directly specified by $\lang$.

Intuitively, $\Delta$ signals that an unknown human preference $\weighta$ is impacting behavior.
If $\Delta$ is 0, then behaviors $\traj$ and $\traj'$ are either the same despite starting in different states or different but in a way conveyed by $\lang$. If $\Delta$ is 1, then $\traj$ and $\traj'$ differ beyond the language specification. 
In the \figref{fig:front_fig} example, the user demonstrations differ despite the specification ``Throw away the can'' not explicitly indicating that they should.
Our hypothesis is that the context change between $\st$ and $\st'$ can reveal the human preference $\weighta$ that resulted in the behavior change in $\traj$ and $\traj'$.

When $\Delta = 1$, we query the LM for potential human preferences $\weighta_i$ that explain the change in behavior for the two scenes, i.e. $\weighta_i \sim \mathrm{LM}_\mathrm{pref}(\Capt(\st), \Capt(\st'), \lang, \Delta=1)$. We denote the set of ``sampled'' preferences $\Weighta_{LM} = \{\weighta_i\}_{i=0}^k$.
The PLGA estimate $\hat\weighta$ should be the most likely in $\Weighta_{LM}$. To generate that, we ask the LM to also assign a normalized probability for how likely it is that $\weighta_i$ is the hidden preference, resulting in a distribution $P(\weighta \mid \st, \st', \lang, \Delta=1)$ with support on $\Weighta_{LM}$. 
In the passive version of PLGA, we simply select $\hat\weighta$ to be the preference in $\Weighta_{LM}$ with the highest probability. %In \secref{sec:lm_active}, we will demonstrate how to additionally actively query the human for preference feedback if the LM is unsure in its hypothesized set.  

\smallskip
\noindent\textbf{Probabilistic Interpretation.} Given $\Delta(\st, \st', \lang)$ as a proxy for ``inexplicable'' change in behavior between states, we assume the LM has a strong prior for modeling $P(\weighta \mid \st, \st', \lang, \Delta(\st, \st', \lang))$. If $\Delta(\st, \st', \lang) = 0$, there is no human preference at play, hence there's no need to update the abstraction. If $\Delta(\st, \st', \lang)=1$, we query the LM for the most likely preferences that explain the change, acting as a sampler, i.e. $\weighta_i \sim P(\weighta \mid \st, \st', \lang, \Delta=1)$, and also for the corresponding probability values, i.e. $ P(\weighta_i \mid \st, \st', \lang, \Delta=1) \forall \weighta_i\in\Weighta_{LM}$. In passive PLGA, we select the preference estimate $\hat\weighta = \argmax_{\weighta_i} P(\weighta_i \mid \st, \st', \lang, \Delta(\st, \st', \lang))$.

% \vspace{-3mm}
\subsection{Querying Preferences with Language}
\label{sec:lm_active}

If the LM model is uncertain about which of the hypothesised preferences $\weighta_i$ is the most likely explanation for the behavior change, PLGA enters an active learning stage where it queries the user directly for the cause of behavior change. This scenario may apply when the human preference cannot be captured by a general LM prior, e.g. ``pick up my favorite object" where the robot is uncertain about what the user's ``favorite object'' may be. In such cases, we expect none of the probability values to stand out. In other words, the entropy of the LM-queried distribution $P(\weighta_i \mid \st, \st', \lang, \Delta=1)$ is high. We propose that when this is the case, the robot should query the human directly for a language description of their preference $\hat\weighta$.

\subsection{Policy Learning with PLGA}
\label{sec:plga_policy}

Once the robot has a preference estimate $\hat\weighta$, our abstraction function is simply $\absfunc^{\mathrm{PLGA}}(\st, \lang, \hat\weighta) = \Capt^{-1}(\mathrm{LM}_\mathrm{abs}(\Capt(\st), \lang, \hat\weighta))$. We can use this to train our policies $\policy_{\hat\weightp}$, similar to LGA:
\begin{equation}
\begin{aligned}
    \mathcal{L}_{\mathrm{PLGA}} = 
   \mathbb{E}_{(\st_t^i,\ac_t^i,\mathbin{\lang^i})\sim \mathcal{D}}
   [||\policy_{\hat{\weightp}}(\mathbin{\textcolor{red}{\absfunc^{\mathrm{PLGA}}}(\st_t^i, \lang^i, \textcolor{red}{\hat\weighta})})-\ac^i_t||^2_2]\enspace.
\end{aligned}
\label{eq_ilga}
\end{equation}
with differences from LGA highlighted in red.

\section{Investigating Passive PLGA as a Prior for General Human Preferences}
\label{sec:sim-exps}

\begin{figure*}[t!]
    \centering
    \includegraphics[width=0.9\textwidth]{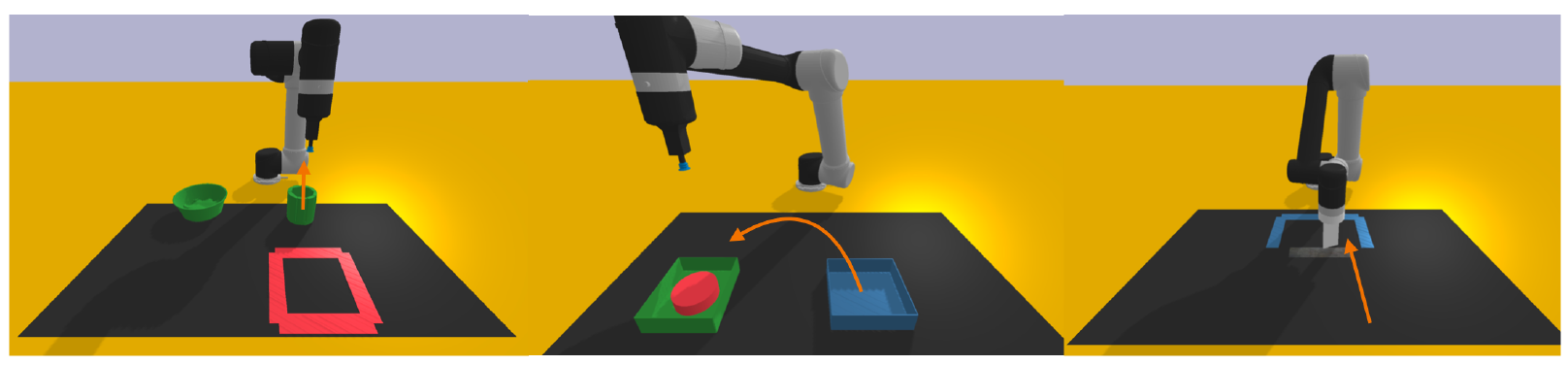}
     \vspace{-3mm}
    \caption{\small We evaluate on three tabletop manipulation tasks: pick, place, and sweep.}
\label{fig:tasks}
\end{figure*}

We begin our evaluation by testing PLGA's ability to leverage the semantic priors in LMs to generate human preferences that explain changes in behavior.
We first conduct simulated experiments to demonstrate passive PLGA in cases where the LM should be able to confidently identify the human preference. For cases where the LM may be unsure about the hidden preference, we will test the active component of PLGA with real users in \secref{sec:human-exps}. Here, we present results for nine different scenarios across three different tasks.

\smallskip
\noindent\textbf{Environment.} We generate a series of robotic control manipulation tasks from the simulated environment VIMA \cite{jiang2022vima} (\figref{fig:tasks}). VIMA is a vision-based simulator where a UR5 arm is tasked with manipulating a specified target object into a desired goal configuration. Observations are top-down RGB images of the manipulation space and actions are continuous pick and place poses each consisting of a 2D coordinate and a rotation expressed as a quaternion. We modify the VIMA feature space to contain up to 48 potential objects (e.g. bowl) and 17 colors/textures (e.g. glass) (see list in Appendix).

Following standard LGA, we implement a captioner module that extracts the feature set $\feat$ from the original RGB observation. This captioner uses a ground truth segmentation mask and labels it with text descriptions of objects and their properties (texture, object ID, etc.). Our PLGA algorithm constructs the task-relevant feature subset $\hat{\feat}$ using GPT4~\cite{openai2023gpt4} as the LM. We query the LM by providing a language utterance, description of the scene, estimated preference, and a target feature to evaluate (the full prompt can be seen in the Appendix). The LM returns a binary response indicating whether that feature should be included in the preference-conditioned abstraction $\hat{\feat}$. Finally, we convert $\hat{\feat}$ to $\hat{\st}$, a binary pixel mask over the robot observation where all identified task-relevant features are represented as ones (otherwise zero).

Our algorithm requires finding trajectory pairs in the demonstration set where the language specification can't explain the behavior change. To generate them, we randomly sample trajectory pairs from $\mathcal{D}$, compute their Euclidean distance and their corresponding preference-free abstractions $\hat\st = \absfunc^{\mathrm{LGA}}(\st, \lang)$ and $\hat\st' = \absfunc^{\mathrm{LGA}}(\st', \lang)$, and check for pairs that are more than $\kappa$ distance apart while mapping to the same abstraction $\hat\st=\hat\st'$. In our experiments, we found $\kappa > 0.2$ was a good metric for differentiating trajectories.

\smallskip
\noindent\textbf{Tasks.}
We investigate three tasks that arise in the context of personal robotics: 1) \textit{pick up the [target]}, 2) \textit{place grasped object on the [target]}, and 3) \textit{sweep object 1 into object 2} [while avoiding potential obstacle] (brackets denote objects the user may have a preference distribution over). For each task, we test three possible (unspecified) human preferences that may impact the desired abstraction.

\begin{enumerate}
    \item For \textit{pick}: 1) a (\textit{ripe}) tomato, 2) a (\textit{container}) to put food in, 3) a (\textit{dry}) cereal bowl (parentheses denote the hidden preference). The robot must determine the correct target object given behavioral context (e.g. \textit{is a green tomato a target pick object?}).
    \item For \textit{place}: 4) a (\textit{non-electronic}) object such as pan, 5) a (\textit{stable}) surface such as coaster, 6) a (\textit{desired content}) container such as recycling or trash. For these tasks, the robot must determine the correct target for the held object to be placed on/in (e.g. \textit{is a laptop a valid place location?});
    \item For \textit{sweep}: 7) a \textit{hot} object such as stove, 8) a \textit{sweepable} object such as rug, and 9) a \textit{sharp} object such as knife. For these tasks, the robot must assess whether objects are potential obstacles to be avoided before executing a sweep motion (e.g. \textit{is a red stove an object to be avoided?)}.
\end{enumerate}

Preferences are instantiated as a distribution over possible object types and colors in the task. These may include preferred pick objects (e.g. red or dark red tomatoes for \textit{ripe}, but not green), preferred place objects (e.g. container or bin for \textit{non-electronic} but not laptop), and avoid obstacles (e.g. a knife for \textit{sharp} but not flower). These are selected to illustrate diversity in preferences that PLGA can infer using strong semantic priors. For each task, the language specification is given without mentioning the preference (e.g. ``Sweep the food into the sink''). PLGA therefore must infer the hidden preference from behavioral context (e.g. \textit{avoid hot objects}). Here we assume there is a generic \emph{but unspecified} preference for each scenario (e.g. users generally prefer to avoid hot objects).

For each preference-task pair, we generate a dataset $\mathcal{D}$ via an oracle demonstrator consisting of 20 demonstrations: 10 expressing behavior when the tested feature is present in the scene and 10 when the tested feature is not (e.g. 10 trajectories of the sweeping food around the stove if the stove is hot, and 10 where sweeping food across the stove otherwise). Target objects are randomly sampled from one of three discretized locations. To create additional complexity, we additionally sample a distractor object that is unrelated to the preference (e.g. a \textit{flower} along with a \textit{stove}).

\smallskip
\noindent\textbf{Manipulated Variables.}
We test PLGA's ability to construct good preference-conditioned abstractions for each task using the LM priors alone. We compare the resulting policies trained via PLGA against two baselines: GCBC (learned directly from raw states and the specified language utterance as per \equref{eq_gcbc}) and LGA (learned from state abstractions constructed via querying $\phi$ against the language utterance alone as per \equref{eq_lga}). We implement GCBC as a goal-conditioned CNN architecture that independently processes language input $\lang$ into an embedding via BERT \cite{devlin2018bert} and the RGB image into an embedding via a CNN, then concatenates the outputs for action prediction via a MLP. We implement LGA and PLGA as the same CNN architecture processing the state abstraction only.

\begin{figure}
    \centering
    \includegraphics[width=0.48\textwidth]{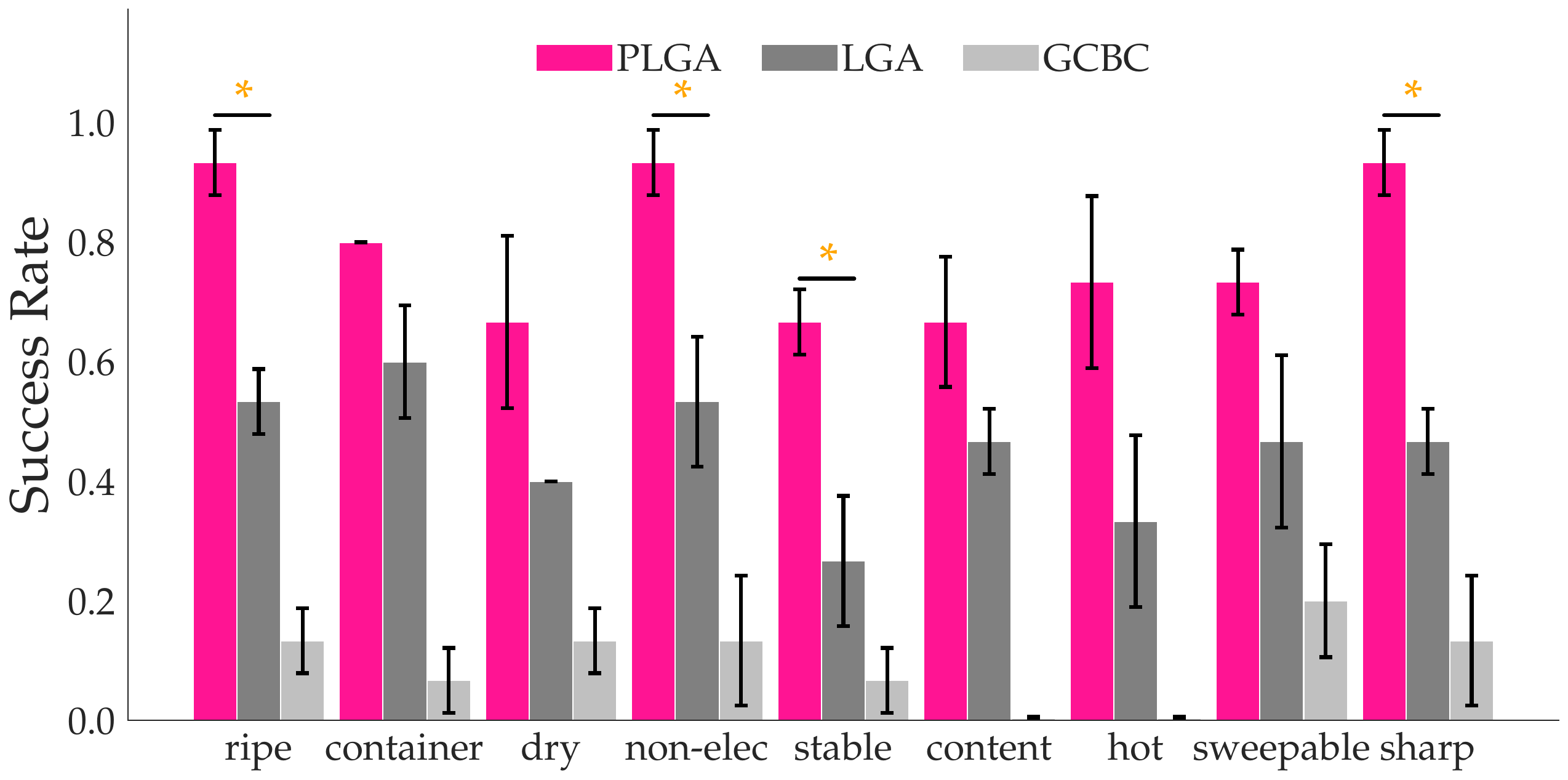}
    \caption{Policy success rate (with standard error) on simulated experiments. PLGA outperforms both LGA and GCBC on task performance, showing better preference-conditioned abstraction construction on downstream task learning.}
    \label{fig:h1}
    % \vspace{-2.5em}
\end{figure}

\smallskip
\noindent\textbf{Dependent Measures.}
We evaluate success as an executed action via a pick/place/sweep of the target object within radius $\alpha$ of the goal.
For these tasks, we constructed a ground truth test distribution reflective of the human preference. We manipulate the training and test distribution such that only a subset of the true preference distribution (e.g. red tomatoes) are seen at training. We evaluate performance via success rate of the learned policies on 5 states sampled from the full test distribution during test.

\smallskip
\noindent\textbf{Hypothesis H1:}
Using information about changes in behavior (PLGA) leads to state abstractions better able to generalize policy learning to preference-conditioned test tasks than abstractions based on language alone (LGA) or no abstractions (GCBC).

\smallskip
\noindent\textbf{Analysis.}
To compare performance, we show in \figref{fig:h1} the policy success rates on test scenes for each task. These results illustrate a trend for better PLGA performance compared to baselines (significant for four tasks with a one-sided t-tests $p<0.05$). 

Overall, this illustrates a trend for better PLGA performance than baselines, supporting the notion that preference-conditioned abstractions enable better generalizable learning. However, one-sided t-tests confirm statistical significance only for four of the tasks. The other tasks display high variance at times in the result, indicating that more trials may be necessary to determine significance. Nevertheless, the qualitative trend softly supports \textbf{H1}.

\section{Investigating Active PLGA for Learning User-Specific Preferences}
\label{sec:human-exps}

In \secref{sec:sim-exps} we tested PLGA's ability to construct \emph{generic} preference-conditioned abstractions using only the LM's priors. We now test its ability to construct abstractions when the preferences are more personalized, meaning the LM may not be entirely sure about its sampled hypotheses $\Weighta_{LM}$. We study the active component of PLGA with a user study to test the ability of PLGA to recognize uncertainty about a preference estimation, causing it to query for the human preference and update its abstraction model accordingly. 

\subsection{Experimental Setup}

\smallskip
\noindent\textbf{Tasks.}
We now construct a new scenario for each task.

\begin{enumerate}
    \item For \textit{pick}: a (\textit{favorite food});
    \item For \textit{place}: a (\textit{preferred dish}) for setting food on;
    \item For \textit{sweep}: a (\textit{specific type of object}) to avoid.
\end{enumerate}

These tasks are now intended to study 1) PLGA's ability to measure uncertainty over the LM's inferred preferences, or in other words, know when it does not know the answer and ask for help and 2) PLGA's ability to update its abstraction generation process given a user-specified preference in natural language.

\begin{wrapfigure}{r}{0.19\textwidth}
\centering
\includegraphics[width=0.19\textwidth]{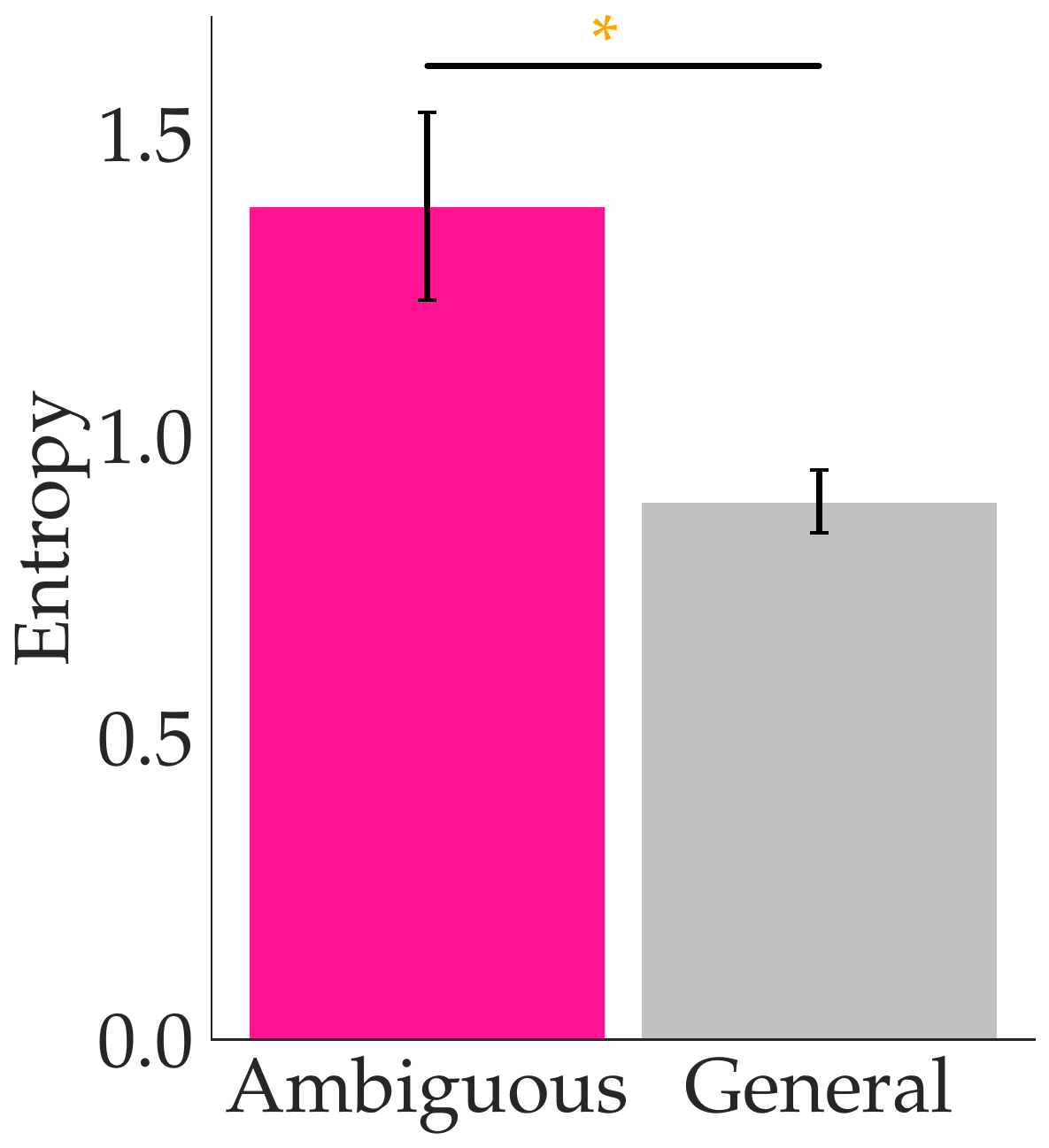}\\
\caption{Entropy values show PLGA can model its own uncertainty under preference ambiguity.}
 \label{fig:entropy}
\end{wrapfigure}

\smallskip
\noindent\textbf{Sanity Check.}
Before investigating PLGA's active querying of human preferences, we first conduct a sanity check to ensure the measured entropy of the resulting LM preference probability is indeed higher (indicating uncertainty) for these tasks vis-a-vis those less ambiguously defined in the previous section. We perform the same LM query as before (e.g. where the LM is tasked with inferring a hidden \textit{favorite food} from $\Delta$). As shown in \figref{fig:entropy}, we do see larger uncertainty for tasks containing more ambiguous preferences, and a one sided t-test ($t(10)=-3.49, p=0.005$) confirms this observation. Based on these results, we found $\epsilon = 1.0$ to be a good entropy threshold for measuring uncertainty.

\smallskip
\noindent\textbf{Study Design.} We conducted a computer-based in-person user study where participants were shown a text description of the task, and asked to give a general preference specified in natural language.

The study is split into three phases: familiarization, scenario generation, and preference querying. During familiarization, we introduce the user to the task context, the simulation interface, and full feature list that is available in the environment. We then show them an example task and text abstraction $\hat{\phi}$. In scenario generation, we introduce six scenarios (two per task), where we describe a background story for each user (e.g. \textit{you are about to have guests over for dinner} or \textit{you now need to figure out how to store food}). This was intended to elicit a natural preference for how each scenario would be interpreted that invoked different downstream preference-conditioned abstractions (e.g. \textit{plate} and \textit{bowl} may be more relevant for the first scenario, while \textit{container} and \textit{box} might be more relevant for the second). In preference querying, we then ask the user to specify, in language, their explicit preferences for the task as our preference query. This preference query is then used by PLGA to explicitly update its abstraction-generation.

\smallskip
\noindent\textbf{Participants.} We recruited 12 participants (50\% male, aged 18-29) from the greater community. We paid participants \$30 for participation. Our study passed institutional IRB review.

\subsection{Subjective Results: PLGA Enables More Natural and Easy User Interaction}
\label{sec:study_subjective}

\begin{figure}
    \centering
    \includegraphics[width=0.5\textwidth]{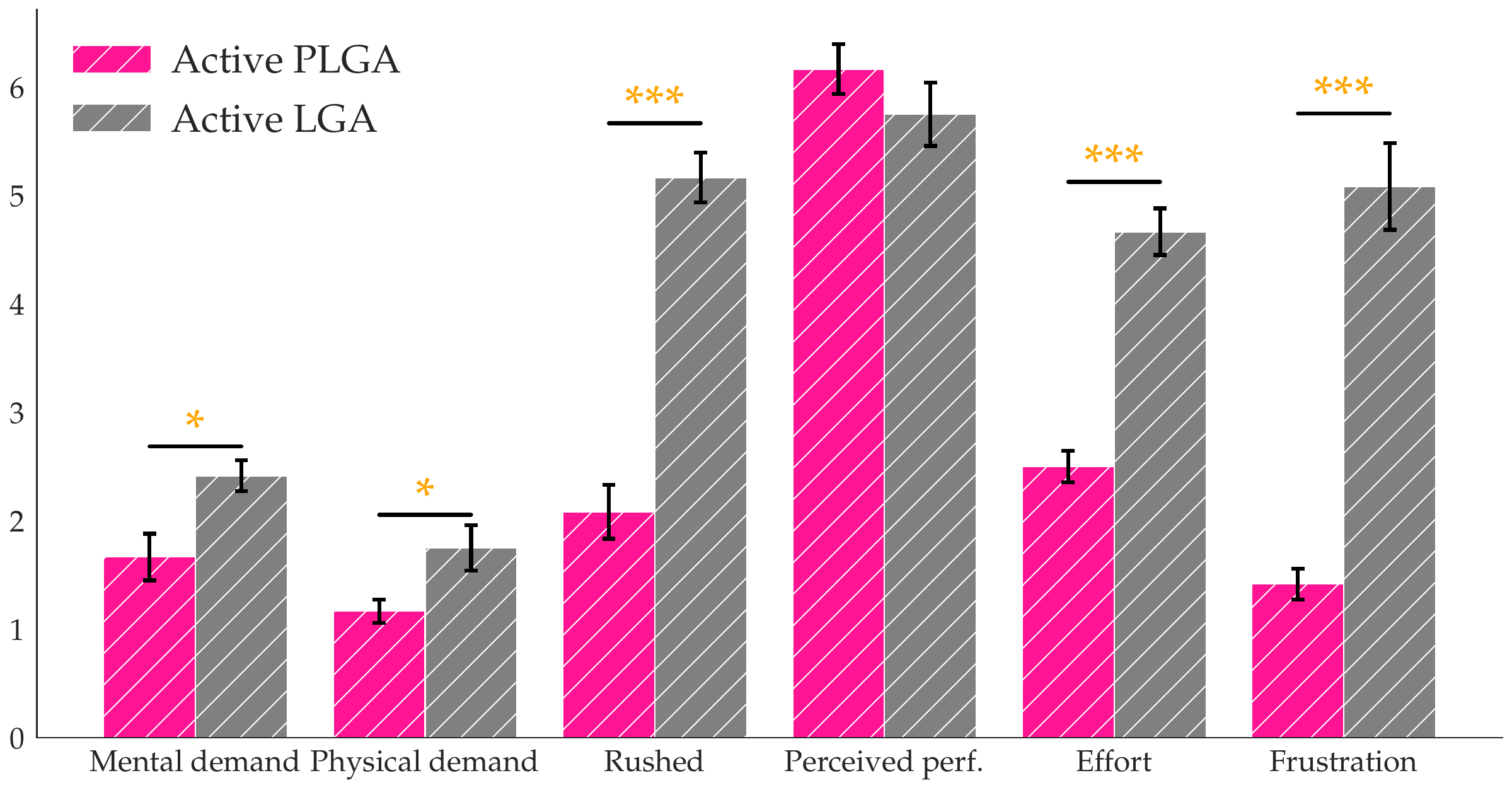}
    \caption{User study interaction results (lower is better for all but perceived performance). The interaction experience with Active PLGA is rated more favorably by users than with Active LGA.}
    \label{fig:h2}
     \vspace{-2mm}
\end{figure}

We first tested if users can easily and effortlessly specify individualized preferences via natural language to the model in a manner that is less burdensome and frustrating than baseline human-in-the-loop abstraction construction methods.

\smallskip
\noindent\textbf{Manipulated Variables.}
We are interested in comparing the user experience of PLGA vs. a baseline human-in-the-loop abstraction method. The baseline we select is the active version of LGA where users are first presented with an LM's best guess of the correct abstraction list (without explicitly modeling preference), and then asked to refine the resulting representation via a text-based interface. We implemented this baseline as an additional condition in our user study. In the active LGA condition, the preference querying phase is instead replaced with an explicit abstraction querying phase, where the user is tasked with specifying, in text, the feature list $\hat{\phi}$ that contains all task-relevant aspects for their preferences in each task. We provide a full list of environment features for easy access. We counterbalance conditions and record qualitative task experience post-conclusion of both conditions.

\smallskip
\noindent\textbf{Dependent Measures.}
For measuring interaction experience, we administered the subjective 7-point Likert Scale survey, inspired by the NASA-TLX \cite{HART1988139}. We presented the survey after the user completed both conditions, and recorded responses for each. 

\smallskip
\noindent\textbf{Hypothesis H2:}
Describing a language preference (Active PLGA) is a more natural and less effortful user interaction experience than manually filtering relevant abstraction features (Active LGA).

\smallskip
\noindent\textbf{Analysis.} \figref{fig:h2} illustrates our subjective user study results with the NASA TLX scores aggregated across participants. We additionally ran paired t-tests with significance level $\alpha=0.05$, marked with orange asterisks. We see that users found PLGA to be significantly less mentally ($t(11)=-2.46, p<0.05$) and physically demanding ($t(11)=-2.54, p<0.05$), and the results are even more pronounced for feeling rushed ($t(11)=-7.40, p<0.001$), frustrated ($t(11)=-8.48, p<0.001$), or expending a great deal of effort ($t(11)=-8.99, p<0.001$). Meanwhile, we found no statistically significant difference in perceived performance ($t(11)=1.60, p=0.14$), suggesting that Active PLGA offers a more natural and effortless interaction experience than Active LGA with no loss in performance quality. Overall, results support our hypothesis \textbf{H2}.

The result is not surprising -- after all, it is to be expected that giving a natural language utterance is an easier experience than inspecting a list of features and selecting the right subset. However, we wanted to verify that users overall find it easy to explicate their preference in words, and that training the robot this way does not decrease their perception of its performance. From this point of view, the results are positive and even encouraging for future research using natural language to explicate human preferences.

\subsection{Objective Results: Active PLGA Successfully Learns from Human Preference Queries}

\begin{figure}[t!]
    \centering
    \includegraphics[width=0.48\textwidth]{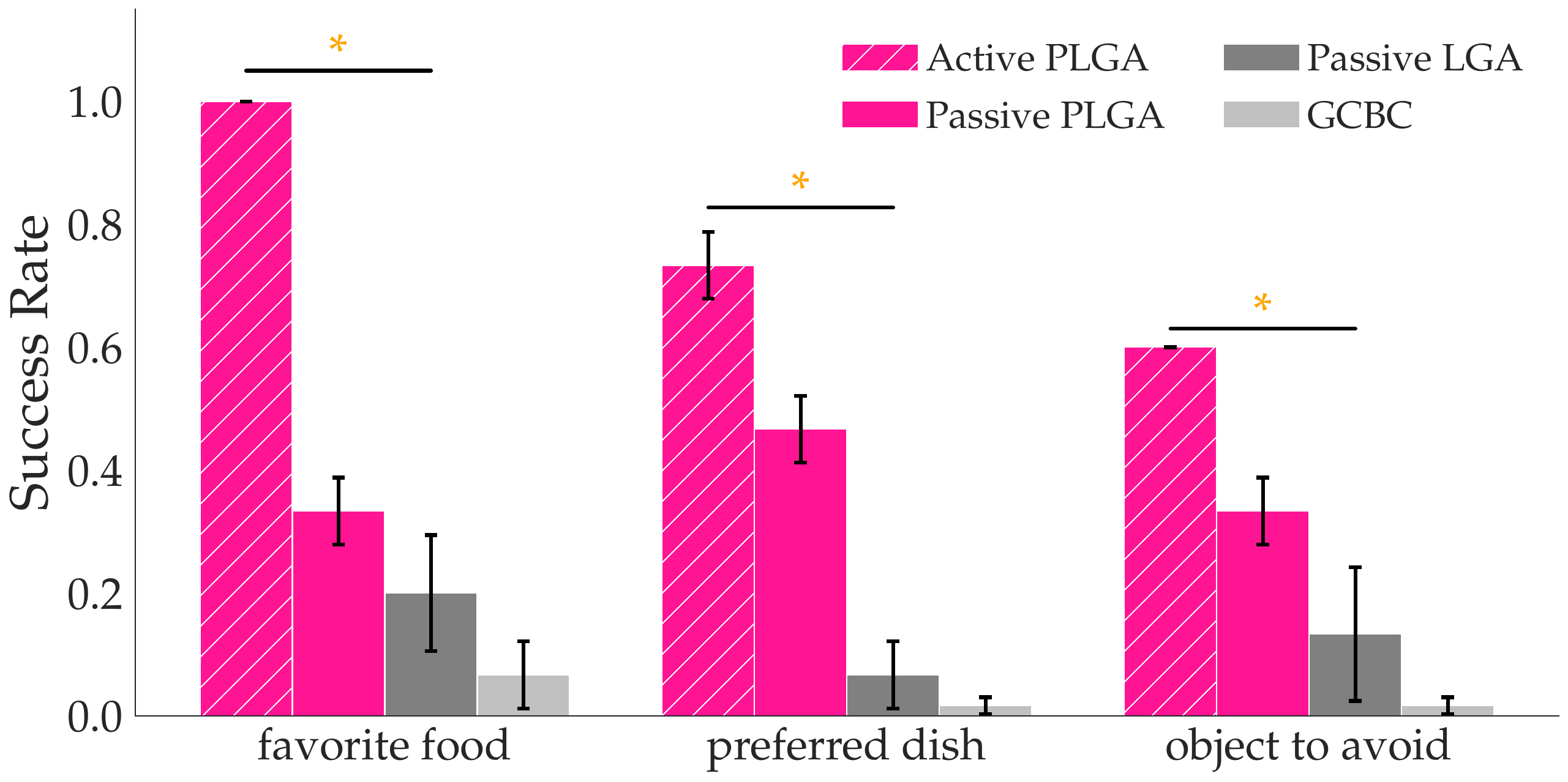}
    \caption{Learned policy success rates for tasks with ground truth preference specified by user study participants. PLGA (active) outperforms PLGA (passive), LGA (passive), and GCBC on task performance, demonstrating an ability to flexibly incorporate natural language human preferences into abstraction construction.}
    \label{fig:h3}
\end{figure}

Now that we have established active PLGA enables a more natural and less effortful user interaction, we measure whether querying users for their preference in natural language results in good preference-conditioned abstractions as compared to baselines.

\smallskip
\noindent\textbf{Manipulated Variables.}
We compare the performance of active PLGA to non-interactive abstraction construction algorithms: Passive PLGA (where the LM did not explicitly query the human for their preference and instead used its best estimate $\hat\weighta\in\Weighta_{LM}$), Passive LGA (where the LM builds an abstraction without explicitly modeling preference), and GCBC. We would like the comparison to validate the importance of identifying when the LM is unsure in its hypotheses and asking the human, when compared to taking its best guess (Passive PLGA), not reasoning about preferences at all (Passive LGA), or not even using state abstractions (GCBC).

\smallskip
\noindent\textbf{Dependent Measures.}
For measuring downstream task success, we report the same success rate as in \secref{sec:sim-exps}. Note, instead of assuming ground truth test distributions constructed by the experimenters, we now assume the abstractions explicitly specified by the human manually during the Active LGA querying in \secref{sec:study_subjective} \textit{are} the ground truth test distributions by which to evaluate. This is a reasonable assumption considering previous work \cite{peng2023diagnosis,bullard2018human,cakmak2012designing} has demonstrated the ability of humans to perform task-specific feature selection to their individualized preferences.

\smallskip
\noindent\textbf{Hypothesis H3:}
Abstractions learned with human preference queries (Active PLGA) result in better performing policies compared to passive methods (Passive PLGA, Passive LGA, GCBC).

\smallskip
\noindent\textbf{Analysis.} \figref{fig:h3} shows that active PLGA outperforms other passive baselines in learning good preference-conditioned abstractions from human queries in natural language, supporting \textbf{H3}. We further confirmed this by running one-sided t-tests (marked with orange asterisks) between Active PLGA and Passive LGA, our strongest competing baseline, confirming significance at $p<0.05$.
This illustrates the ability of PLGA to integrate information queried from the user meaningfully in constructing state abstractions. 
Moreover, while every method has its natural user effort vs. information gain tradeoff, PLGA's ability to query seamlessly for natural human feedback while reducing user frustration and effort is an exciting testament to the value of strong priors for preference learning.

\section{Investigating PLGA on a Spot Robot}
\label{sec:robot-exps}

We demonstrate the real world abstraction construction utility of PLGA on a Spot robot\footnote{Our Spot's name is Moana.} performing mobile manipulation tasks.

\smallskip
\noindent\textbf{Robotic Platform.}
Spot is a mobile manipulation legged robot equipped with six RGB-D cameras (one in gripper, two in front, one on each side, one in back), each producing an observation of size 480x640. We only use observations taken from the front camera.

\smallskip
\noindent\textbf{Tasks and Data Collection.}
We collected demonstrations of a human teleoperating the robot while performing two mobile manipulation tasks with household objects: \textit{place the drink in the bin} and \textit{throw away the can}. The manipulation action space consists of the following three actions along with their parameters: (\textit{xy, grasp}), (\textit{xy, move}), (\textit{drop}) while the navigation action space consists of a SE(3) group denoting robot waypoints\footnote{For ease of data generation, we perform imitation learning over the trajectory rather than each state (i.e. predict a sequence of actions from an initial observation).}. 
For \textit{place the drink}, the robot is tasked with bringing an already-grasped soda can to a specified location and dropping it into a trash can. We assume the user has a preference for avoiding electronics in the way, otherwise taking the shortest path. 
For \textit{throw away}, the robot is tasked with picking up a drink on a table, bringing it to a correct bin (either recycling or trash), and successfully dropping the drink into the bin. We assume the user has a preference for placing cans in a recycling bin if one is available, and otherwise placing them in the trash.  Both tasks include possible distractors like drills and brushes. 

For \textit{place the drink}, we generate demonstrations of the robot placing a soda can into the recycling if available, otherwise trash. At test time, we evaluate the robot on the scenarios with a water bottle instead. For \textit{throw away the can}, we generate demonstrations of the robot walking directly to the trash can when a shirt is on the ground, but avoiding the drill when it is present. At test time, we evaluate the robot on two new scenes: a laptop (to avoid) and pants (walk across). While the robot sees a trajectory of a user avoiding a drill during train, it is not exposed to laptops prior to test.

\smallskip
\noindent\textbf{Training and Test Procedure.}
We first extract a segmented image from the observations using Segment Anything \citep{kirillov2023segment} and captioner Dedic \citep{zhou2022detecting} to perform a check for behavior $\Delta$ (e.g. is the robot taking a different trajectory when a laptop is present in the scene vs. shorts). If the answer is yes, we instantiate the full PLGA pipeline. First, we perform a preference query to the LM with the initial two scenes and task description; next, we use this preference to query the LM to construct a preference-conditioned abstraction; lastly, we map this abstraction back into the observation dimension.

\smallskip
\noindent\textbf{Takeaway.}
PLGA produced policies capable of successfully completing both tasks consistently, even when faced with new distractor objects, target object colors, or unseen linguistic specifications. Excitingly, we were able to observe non-trivial generalization capabilities, particularly in the avoid task (the robot successfully learned to avoid laptops from only seeing a demonstration of avoiding a drill). The failures we did observe were largely due to captioning errors (e.g. the segmentation model detected the object but was unable to produce a good text description). Our demonstration of PLGA on real robotic hardware indicates an exciting future in using LMs to help generate preference-conditioned state abstractions.
\section{Related Work}
\label{sec:related-work}

\noindent\textbf{Learning from Human Input.}
Existing frameworks for interactive querying for downstream learning, like TAMER~\citep{knox2008tamer} and COACH~\citep{macglashan2017interactive}, use human feedback to train policies, but are restricted to binary or scalar labeled rewards~\cite{abel2017agent,zhang2019leveraging}. Another line of work looks at learning from human preferences, often by asking them to compare or rank trajectory snippets \citep{christiano2017deep,brown2020better}. There are also works that actively learn from human teachers, where the emphasis is on generating actions or queries that are maximally informative for the human to label \citep{bobu2022learning,chao2010transparent}. Unfortunately, these approaches all are limited by the fact that the feedback asked of the human is overfit to specific failures or desired data points, and rarely scale well relative to human time or effort \cite{bobu2023aligning}. 

\smallskip
\noindent\textbf{Language Models for Human Preferences.}
LMs are increasingly being used for personalized applications. Prior work has explored using LMs for recommendation systems \cite{Wu2023ASO,Ji2023GenRecLL,Lyu2023LLMRecPR,mao-etal-2023-unitrec,10.1145/3534678.3539382}, user-specific chatbots \cite{zhang-etal-2018-personalizing,10.1145/3404835.3462828,li-etal-2016-persona,ijcai2018p595,ijcai2019p721,zhong-etal-2022-less}, or even sorting household objects according to personal preferences \cite{wu2023tidybot}. 

A range of techniques have been introduced to specify human preferences and inject them into LMs.
With the popularization of prompting-based techniques, users simply have to write a textual description (called a \textit{prompt}) specifying their preferred task and condition LMs on this prompt to induce their desired behavior \cite{gpt3}.
In order to encourage LMs to produce outputs in line with users' preferences, recent work has explored techniques such as instruction-tuning \cite{InstructGPT,honovich-etal-2023-unnatural,wang-etal-2023-self-instruct,chung2022scaling,zhang2023instruction} and reinforcement learning from human feedback (RLHF) \cite{Bai2022TrainingAH,ziegler2020finetuning,NIPS2017_d5e2c0ad,NEURIPS2020_1f89885d,ganguli2022red}.

Furthermore, having been pre-trained on large corpora of human-generated text \cite{2019t5},
LMs often possess sensible priors over ``typical''\footnote{It is worth noting that text scraped from the internet, which constitutes the bulk of what today's LMs are trained on, is biased and does not capture a representative sample of human preferences globally.}) human preferences and behaviors \cite{li2023lampp,gpt3,Zhou2019EvaluatingCI}.
Because of this, LMs have at times even been used as \textit{simulations} of humans \cite{aher2023using,DILLION2023597,argyle_busby_fulda_gubler_rytting_wingate_2023}. As part of prompting, LMs must implicitly perform language understanding on human-written prompts to infer their preferences. However, LMs have also been used to \textit{explicitly} infer human preferences from linguistic specifications. For example, recent work has examined reward learning using LMs \cite{lin-etal-2022-inferring,kwon2023reward}.

\smallskip
\noindent\textbf{Language Models in Robotics.}
LMs hold commonsense knowledge about object properties, functions, and their relevance to various tasks. This is why many recent works have explored using LMs to output plans directly, i.e. generate primitives or high-level action sequences~\citep{sharma-etal-2022-skill,ahn2022can,huang2022language,huang2022inner}.
These approaches use priors embedded in LMs to produce better \textit{instruction following} models, or in other words, better compose base skills to generate more complex behavior \citep{zeng2023socratic,li2023lampp,ahn2022can,wang2023voyager}. In contrast, we use LM priors to learn \textit{it} preferences over relevant features. Recent work \citep{peng2023lga} has also proposed to use LMs to perform \textit{state abstraction} for learning better skills \textit{from scratch}, instead leveraging the LM's priors to identify task-relevant features for state abstraction construction.
\section{Discussion}
\label{sec:conclusion}

We presented PLGA, a framework for learning preference conditioned state abstractions from language and demonstration information.
Particularly, we focused on settings where the language task specification does not list everything the human cares about. 
We introduced LM preference queries for inferring user preferences present in demonstrations directly from LM priors.
Our simulated experiments, user study, and Spot robot demos illustrate that natural language can be a convenient vehicle to communicate hidden preferences for constructing state abstractions, and those abstractions result in improved downstream task performance. Although we demonstrated PLGA's real-world applicability in home manipulation tasks, we are excited about future opportunities in shared autonomy tasks (where the human may have a preference for which aspects of the task the robot assists with), or autonomous driving (where users have a preference for what objects to avoid).

\smallskip
\noindent\textbf{Limitations and Future Work.}
In our work, we assumed we had no further information regarding %the user's 
differences in user behavior beyond the initial states that induced these behaviors.
However, we do not use the information about \textit{how} exactly user behavior changed.
A natural direction would be to extend PLGA's preference query abilities to user trajectories, where richer features, like obstacle avoidance distance, can be explored. Such a path would open more meaningful opportunities for grounding natural language to the language of human behavior.

Moreover, while we focused here on using language priors to construct state abstractions for imitation learning, a natural parallel would be to explore this framework in the context of rewards, where rich semantic priors could be extremely meaningful to few-shot downstream learning from demonstrations. 
Furthermore, our algorithm is not designed to be iterative, which means that there is no opportunity for continual preference learning after repeated exposure to different interactions.
However, there are many trajectory-based features that arise in the context of robotics that would require more text-based motion information regarding user actions that we currently do not have.

Lastly, while we broached the subject of active preference elicitation, we did not conduct a deep dive into meaningful ways to interact with the user when trying to learn their preference (opting instead to query them directly if uncertain). Future work can explore different ways of performing preference elicitation with language models, including iterative approaches that perform sequential updates to the reward or preference model. 
\section{Acknowledgements}

We thank Jacob Andreas and Jessica Hodgins for feedback and brainstorming of ideas, as well as the Boston Dynamics AI Institute for providing hardware resources used in this paper. We additionally thank members of the MIT Interactive Robotics Group and Language and Intelligence Group for helpful discussions. Andi Peng is supported by the NSF Graduate Research Fellowship and Open Philanthropy. Belinda Li and Theodore Sumers are supported by National Defense Science and Engineering Graduate Fellowships. Ilia Sucholutsky is supported by a NSERC Fellowship (567554-2022).

\newpage

\bibliographystyle{ACM-Reference-Format}
\balance
\bibliography{references}

\clearpage
\appendix
\balance
\section{Appendix}
\label{sec:appendix}

\subsection{Full Prompt}
ChatGPT models (including GPT4) can take in both system prompts and user prompts. We split our prompt into these two parts.

\textbf{Preference Query.}
System prompt where \{scene\_intersection\} is replaced by the list of all similar features between two scenes and \{scene1\_minus\_scene2\} and \{scene2\_minus\_scene1\} are the lists of scene differences.

\begin{quote}
    There are two scenes. The user takes a different trajectory in the first scene vs. the second. 

    The first and second scene both have the following features:
    \{scene\_intersection\}\\
    The first and second scene differ on the following:\\
    First scene-
    \{scene1\_difference\}\\
    Second scene-
    \{scene2\_difference\}\\

    What are the most likely high-level preferences to have caused the difference in the user's behavior and why? The user took different trajectories in the two scenes. Please give a list of brief preferences (with only one reason) and assign a confidence score to each answer, in the format [["answer", score], ["answer", score], ...]. Please ensure all scores sum up to 1.
\end{quote}

\textbf{Abstraction Query.}
System prompt where \{object\_list\} is replaced by the list of all object types in the environment and \{object\_colors\} by the list of all colors and textures:
\begin{quote}
    You are interfacing with a robotics environment that has a robotic arm learning to manipulate objects based on some linguistic command (e.g. ``pick up the red bowl''). At each interaction, the researcher will specify the command that you need to teach the robot. In order to teach the robot, you will need to help design the training distribution by specifying what properties task-relevant objects can have based on the given command. Objects in this environment have two properties: object type, object color.  Any object type can be paired with any color, but an object can only take on exactly one object type and exactly one color.\\
Object types:\\
\{object\_list\}\\
Object colors:\\
\{object\_colors\}
\end{quote}
User prompt where \{rule\} is replaced by one of the task prompts listed above, \{group\} is replaced by ``object color'' or ``object type'', and \{candidate\} is replaced by each candidate object color or type that we would like the LM to evaluate:
\begin{quote}
    The command is ``\{rule\}''. In an instantiation of the environment that contains only some subset of the object types and colors, could the target object have \{group\} ``\{candidate\}''? Think step-by-step and then finish with a new line that says ``Final answer:'' followed by ``yes'' or ``no''.
\end{quote}

\subsection{Task Details}

\textbf{Pick:}

\textit{ripe tomato}:
\begin{itemize}
    \item Task description: \textit{Bring me a tomato.}
    \item True distribution: \{objects: tomato\}, \{textures: red, dark red\}
\end{itemize}

\textit{food container}:
\begin{itemize}
    \item Task description: \textit{Bring me something to put food in.}
    \item True distribution: \{objects: bowl, container, box\}, \{textures: ALL\}
\end{itemize}

\textit{dry cereal bowl}:
\begin{itemize}
    \item Task description: \textit{Bring me a cereal bowl}
    \item True distribution: \{objects: bowl, drying rack, drying towel, drying cloth\}, \{textures: ALL\}
\end{itemize}

\textbf{Place:}

\textit{non-electronic}:
\begin{itemize}
    \item Task description: \textit{Put down my mug.}
    \item True distribution: \{objects: ALL \\ iPad, laptop, phone\}, \{textures: All\}
\end{itemize}

\textit{stable surface}:
\begin{itemize}
    \item Task description: \textit{Put down the pan.}
    \item True distribution: \{objects: pan, coaster, pallet\}, \{textures: ALL\}
\end{itemize}

\textit{desired content}:
\begin{itemize}
    \item Task description: \textit{Put away my food.}
    \item True distribution: \{objects: tomato, pepper, peach, apple, container, box\}, \{textures: ALL\}
\end{itemize}

\textbf{Sweep:}

\textit{hot object}:
\begin{itemize}
    \item Task description: \textit{Sweep the food into the sink.}
    \item True distribution: \{objects: food, sink, stove, pan\}, \{textures: red, dark red\}
\end{itemize}

\textit{sweepable}:
\begin{itemize}
    \item Task description: \textit{Sweep the dust into the container.}
    \item True distribution: \{objects: bin, container, floor\}, \{textures: wooden, granite\}
\end{itemize}

\textit{sharp}:
\begin{itemize}
    \item Task description: \textit{Sweep the food into the sink.}
    \item True distribution: \{objects: pepper, peach, apple, sink, knife, sharp block\}, \{textures: ALL\}
\end{itemize}

\end{document}